\documentclass[11pt]{article}
\usepackage{listings}
\usepackage{graphicx}
\usepackage{amsmath}
\usepackage[affil-it]{authblk}
\usepackage{booktabs}
\usepackage{multirow}
\usepackage{array}
\usepackage{tabularx}
\usepackage[numbers]{natbib}
\usepackage{hyperref}

\usepackage[margin=1in]{geometry}

\author[1]{Alessio Sordo$^*$}
\author[1]{Lingxiao Du$^*$}
\author[ ]{Meeka-Hanna Lenisa}
\author[1]{Evgeny Bogdanov}
\author[1]{Maxim Romanovsky}

\affil[1]{Deutsche Bank, Berlin Technology Center, Berlin, Germany}

\date{}

\begin{document}

\title{STELLAR-E: a Synthetic, Tailored, End-to-end LLM Application Rigorous Evaluator}
\maketitle
\def\thefootnote{*}\footnotetext{Both authors contributed equally to this research.}\def\thefootnote{\arabic{footnote}}

\begin{abstract}
  The increasing reliance on Large Language Models (LLMs) across diverse sectors highlights the need for robust domain-specific and language-specific evaluation datasets; however, the collection of such datasets is challenging due to privacy concerns, regulatory restrictions, and the time cost for manual creation. Existing automated benchmarking methods are often limited by relying on pre-existing data, poor scalability, single-domain focus, and lack of multilingual support. We present STELLAR-E --- a fully automated system to generate high-quality synthetic datasets of custom size, using minimal human inputs without depending on existing datasets. The system is structured in two stages: (1) We modify the TGRT Self-Instruct framework to create a synthetic data engine that enables controllable, custom synthetic dataset generation, and (2) an evaluation pipeline incorporating statistical and LLM-based metrics to assess the applicability of the synthetic dataset for LLM-based application evaluations. The synthetic datasets reach an average difference of +5.7\% in terms of LLM-as-a-judge scores against existing language-specific benchmarks, demonstrating comparable quality for comprehensive assessment of big and small LLMs. While real datasets remain slightly more challenging for LLMs especially for smaller models, this work establishes a scalable and domain-adaptable benchmarking framework that supports fair evaluation of LLM applications, offering a faster alternative to manual approaches and enabling high-efficiency automated quality assurance cycles.
\end{abstract}

\section{Introduction}
\label{intro}

The evolution of LLMs has enabled powerful new applications, but their non-deterministic nature makes objective evaluation a critical task for ensuring safety and reliability~\cite{Ouyang_2025}.
Currently, evaluating LLM-based systems relies on comparing model outputs against pre-existing datasets~\cite{chang2023surveyevaluationlargelanguage}.
However, the creation of these datasets still frequently depends on human experts, a process that is slow, expensive, and difficult to scale~\cite{liu2024datasetslargelanguagemodels}.

This dependence on manual annotation makes the evaluation process a significant bottleneck.
Industries, especially in regulated fields like healthcare and finance, require continuous benchmarking to detect biases, regulatory compliance breaches, hallucinations, and performance degradation~\cite{Raza2025}.
For these environments, the ability to customize evaluations for specific domains and languages \textemdash without using sensitive real data \textemdash makes fully synthetic evaluation datasets and automated benchmarks an urgent necessity.

Designing an automated pipeline to generate high-quality customizable synthetic datasets and run evaluations on LLM-based systems requires the use of LLMs themselves, which provide flexible automation beyond template-based and programmatic approaches~\cite{long2024llmsdrivensyntheticdatageneration}.
The use of LLMs for these purposes, however, is not straightforward and may introduce challenges including:
biases from the generator or judge LLM~\cite{panickssery2024llmevaluatorsrecognizefavor, stureborg2024largelanguagemodelsinconsistent};
undetected hallucinations;
the generation of low-complexity data that fails to adequately challenge the model under evaluation.

While some automated generation approaches exist, they have key shortcomings.
Many only augment or anonymize existing data, making them unsuitable for high-regulated environments.
Those that generate fully synthetic data often focus on narrow domains or have poor supervision mechanisms, making the process unreliable.

Furthermore, a critical gap exists in the multilingual domain, as most benchmarks are heavily skewed toward English.
Creating non-English datasets via translation is a common workaround, but it introduces significant issues.
Professional translations are costly, while automated translations often contain artifacts like "translationese"~\cite{Gellerstam1986TranslationeseIS}.
Moreover, these translations frequently ignore cultural specificities, resulting in content that is irrelevant or incorrect for the target culture~\cite{singh2025globalmmluunderstandingaddressing}.
This gap widens the disparity between high-resource languages (like English, Spanish, and German) and low-resource languages (like Swahili, Punjabi, or Vietnamese), which lack sufficient evaluation resources~\cite{liu2024datasetslargelanguagemodels}.

To address these problems, we present STELLAR-E, an LLM-powered pipeline to automatically generate high-quality, multilingual instruction-answer datasets in text format and evaluate multilingual and domain-specific fine-tuned LLMs on them.
The generated instructions don't only demand responses but can also require multi-step reasoning or specific actions, which better assess the ability of LLMs to handle complex tasks.
STELLAR-E is designed with robust, multi-stage supervision to ensure dataset quality and requires minimal human input for fast, customizable evaluation cycles.
The system allows for flexible control over the semantics, language, format, and number of the generated instruction-answer (I\&A) pairs.

This solution consists of a set of blocks, each of them dedicated to handle a different stage of the evaluation pipeline. The blocks are independent, allowing for a robust recovery in case of failures. The key contributions of our work are:

\begin{itemize}
    \item A highly customizable, semantics and language-adjustable instruction-answer generation pipeline, with thorough supervision mechanisms.
    \item Toggleable quality optimizations: Diversity Enhancement (DVE) and Difficulty Enhancement (DFE).
    \item Evaluation system implementation and run of meta-evaluations against popular existing datasets to ensure the effectiveness of the automatic pipeline.
\end{itemize}

\section{Related Work}
\label{related_work}

Traditional LLM evaluation on static, human-crafted datasets is becoming inefficient due to issues like saturation and data contamination~\cite{li2024llmsasjudgescomprehensivesurveyllmbased,chen2025recentadvanceslargelangauge,zhou2023dontmakellmevaluation}.
While emerging dynamic techniques offer an alternative, they often rely on transforming existing expert-created datasets, a costly and time consuming process, and may not offer a complete automated pipeline covering generation, curation (quality filtering and correction), and evaluation (reliable, unbiased scoring)~\citep{chen2025recentadvanceslargelangauge}.
In this context, we analyze end-to-end automated systems that focus on textual data: YourBench~\cite{shashidhar2025yourbencheasycustomevaluation}, OmniEval~\cite{wang2025omnievalomnidirectionalautomaticrag}, and BENCHAGENTS \cite{butt2024benchagentsautomatedbenchmarkcreation};
they are chosen for their complementary approaches.

\textbf{YourBench} Shashidhar et al. \cite{shashidhar2025yourbencheasycustomevaluation} generates question-answer pairs (QAs) grounded in documents.
It ingests user-provided documents, preprocesses them, and generates QAs with an ensemble of generation LLMs to improve diversity and coverage.
The curation consists of an algorithm to validate citations and semantic deduplication of questions.
The evaluation is performed by an ensemble of judge LLMs to enhance reliability.
The overall procedure is well-detailed, but results remain dependent on input document quality and coverage.

\textbf{OmniEval} Wang et al. \cite{wang2025omnievalomnidirectionalautomaticrag} is a RAG benchmark for the financial domain.
It builds a knowledge corpus from which it generates benchmark instances with the assistance of GPT-4 agents.
The authors define five RAG tasks, and a data generation agent outputs QA pairs grounded on document passages.
The curation of data is multi-stage with automated filtering and human annotations on a sampled sub-dataset.
The evaluation benchmarks RAG systems via both rule-based and LLM-based metrics.
OmniEval provides an end-to-end RAG assessment; it is inherently dependent on input documents and targeted to a vertical domain where constructing a high-quality knowledge corpus is feasible.

\textbf{BENCHAGENTS} Butt et al. \cite{butt2024benchagentsautomatedbenchmarkcreation} treats benchmark creation as agent orchestration.
Given a task description, a planning agent proposes a plan that is followed by other agents to generate, validate, and evaluate instances.
Curation is performed by a verification agent that defines and runs code --- which is manually reviewed --- to perform pre-defined quality checks.
This agent-based approach does not rely on document corpora, but its effectiveness depends on the design of the agents and on manual code reviews, which make it unsuitable for the creation of large datasets.
This approach has limited scalability in terms of diversity, because there is no guarantee that the iterations produce a diverse enough dataset, as the outputs are evaluated by an agent one-by-one, but are not post-processed in group by a further agent.

Overall, these papers present well-structured evaluation pipelines.
YourBench and Omnieval both target systems where grounding on input documents is required; this reliance slows down the customization to new domains, especially if documents need to be carefully anonymized \cite{cirillo2025augmentinganonymizeddataai}.
BENCHAGENTS adopts a different paradigm, by forming a multi-agentic pipeline without any reliance on input documents, but it presents a lack of post-processing techniques to guarantee the diversity of questions.

Reducing at minimum the reliance on existing data in benchmarks enables to quickly create and update evaluations for custom LLM-based systems, allowing for fast monitoring in LLMOps CI/CD pipelines \cite{pahune2025transitioning}.
Some of the analyzed benchmarks already provide contributions to this concept: Omnieval presents the idea of guiding the generating LLM on a topic created by another LLM, and BENCHAGENTS has no reliance on input data.

The major gap common to all the analyzed pipelines is the limited scalability in terms of evaluation dataset size.
Firstly, because none of the systems allow to easily specify the number of questions to generate.
Secondly, because in the curation every system just filters out the generated instances that do not comply to certain criteria, without acting on the low-quality instances; this logic can reduce the number of generated QAs.
While Omnieval implements a simple correction mechanism, there is no further evaluation of the corrected items, which could prove the actual correction effectiveness.

\section{Methodology}
\label{methodology}

\begin{figure*}[htbp]
    \begin{center}
        \includegraphics[width=1\linewidth]{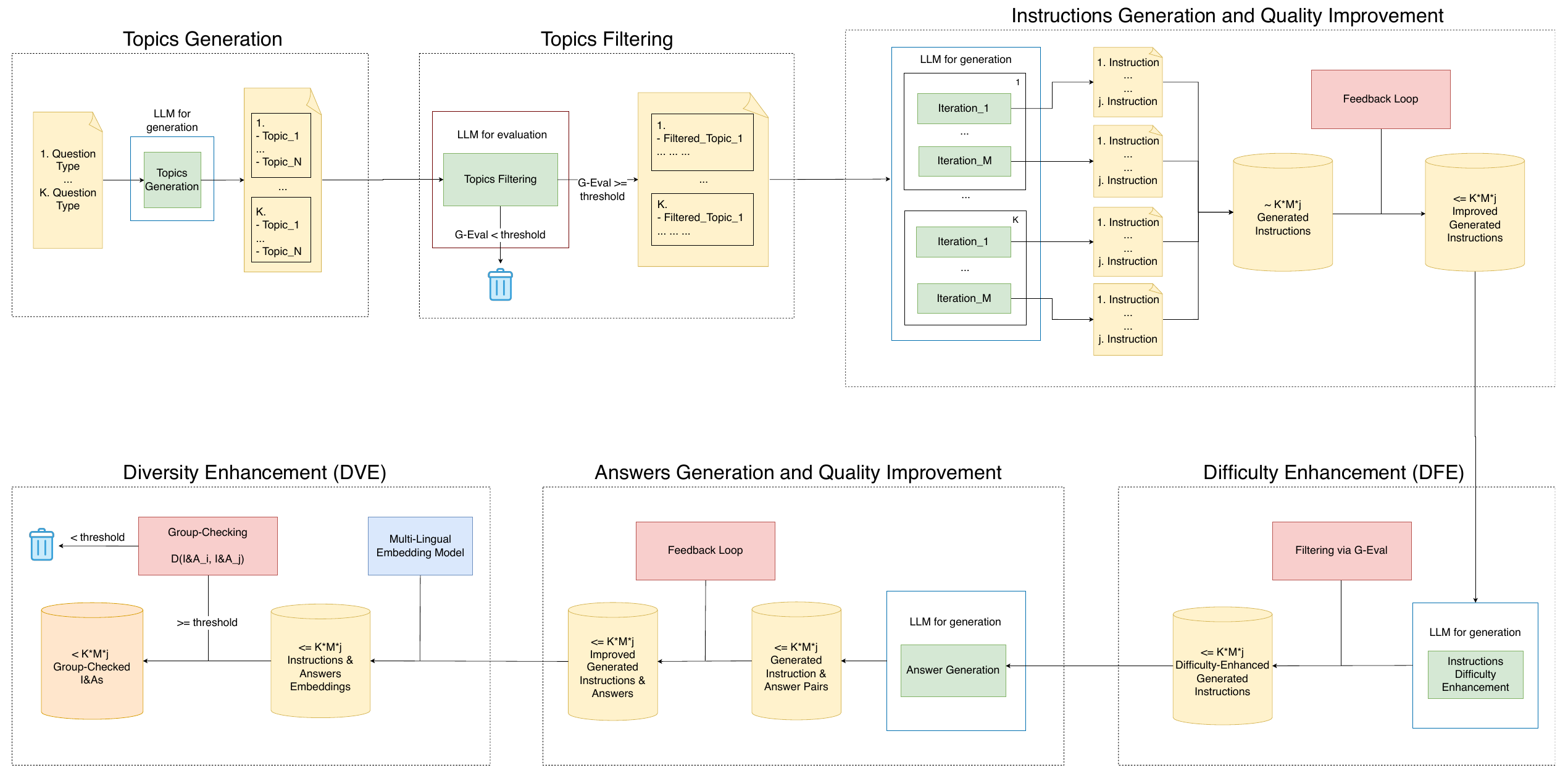}
    \end{center}
    \vspace{-0.1in}
    \caption{Overview of generation pipeline}
    \label{fig:pipeline}
\end{figure*}

Our synthetic instruction and answer generation pipeline is designed to efficiently produce high-quality, diverse, and faithful datasets for LLM training and evaluation. Figure~\ref{fig:pipeline} presents an overview of our process.
The process begins by generating $N$ topics across predefined question types, followed by a topic filtering phase. Next, a random subset of the filtered topics is selected, and for each topic set, $j$ instructions are generated using prompts specifically designed to maximize both diversity and coverage. Subsequent quality improvement and difficulty enhancement steps further refine the instruction set and its corresponding answer set.

We developed and implemented a generation pipeline comprising multiple stages, each dedicated to addressing specific aspects of data generation, filtering, and refinement. The methodology integrates custom implementations, including a tailored G-Eval framework, to enhance the quality and reliability of the generated data. Section~\ref{custom_geval} introduces our custom G-Eval, which is employed throughout various stages of the generation pipeline. This implementation mitigates known limitations of the original G-Eval \cite{liu2023gevalnlgevaluationusing}, such as bias and coverage gaps. Subsequently, Section~\ref{synthetic_data_generation_pipeline} provides a comprehensive description of each stage involved in the generation process.

\subsection{Custom G-Eval}
\label{custom_geval}

For evaluations and meta-evaluations, the G-Eval framework~\cite{liu2023gevalnlgevaluationusing}, based on LLM-as-a-judge with Chain-of-Thought reasoning, is adapted.
Despite the promising results achieved by the original implementation of G-Eval, several limitations have been identified~\cite{stureborg2024largelanguagemodelsinconsistent}.
To address these issues, the original G-Eval methodology is modified in this work by implementing a 1\textendash10 scoring scale for finer granularity, consistently setting the temperature parameter to 0 to enhance generation reliability, and running separate evaluations for each individual parameter/criterion in order to mitigate the anchoring bias.

Furthermore, the selection of the evaluation criteria within G-Eval is task-dependent, as delineated in~\cite{liu2023gevalnlgevaluationusing}.
In this study, the criteria and parameters vary across different stages, which will be described in detail in subsequent sections.
The overall G-Eval score is calculated as the mean of the scores assigned to each criterion.
As LLM-as-a-judge models, we employ Gemini 2.0 Flash 001 in the generation pipeline and Gemini 2.5 Pro in the evaluation phase.

Finally, within the filtering and evaluation stages of our generation pipeline, we establish a threshold $\tau = 8$ (on a 1\textendash10 scale) to denote a satisfactory G-Eval score.
This threshold was selected to minimize the acceptance of borderline outputs.
\subsection{Synthetic Data Generation Pipeline}
\label{synthetic_data_generation_pipeline}

\subsubsection{\textbf{Question Type Definition}}
\label{question_type_definition}

Question Types (QTs) refer to a curated collection of textual elements, each representing a distinct domain of interest.
The selection of question types forms the foundation for the downstream synthetic data generation pipeline.
This idea is inspired by~\cite{sun2023principledrivenselfalignmentlanguagemodels}, where the authors introduce 20 types of adversarial questions \textit{that machine learning models either cannot answer or are likely to answer incorrectly}. These question types are used to facilitate the creation of instruction-answer pairs for LLM alignment, thereby minimizing the need for human intervention. They serve as the inputs to our generation pipeline and are intended to be manually designed by researchers or practitioners to accommodate research objectives or application requirements.

\subsubsection{\textbf{Topic Generation and Filtering}}
\label{topic_generation_and_filtering}

For each question type, an LLM is prompted to generate a diverse list of candidate topics.
A topic is defined as a set of concise noun phrases, each containing no more than three words, generated by an LLM and is closely associated with the specified QT.
This concept is, again, inspired by the work of \cite{sun2023principledrivenselfalignmentlanguagemodels}, who generate topics from a set of adversarial question types.
These generated topics are subsequently evaluated using our custom G-Eval framework, which assesses each topic based on explicit criteria: relatedness to the originating question type and semantic correctness.
Only those topics that exceed predefined quality threshold $\tau$ for both criteria are retained for further consideration.
This ensures that only topics demonstrating a high degree of relatedness and semantic correctness are advanced.
This approach facilitates customization across diverse domains, as empirically demonstrated by \cite{wang2024codeclmaligninglanguagemodels}, who showed that domain-specific tailoring of synthetic data can significantly improve the performance of downstream tasks.

\subsubsection{\textbf{Instruction Generation and Quality Improvement}}
\label{instruction_generation_and_quality_improvement}

\begin{figure}[htbp]
    \begin{center}
        \includegraphics[width=1\linewidth]{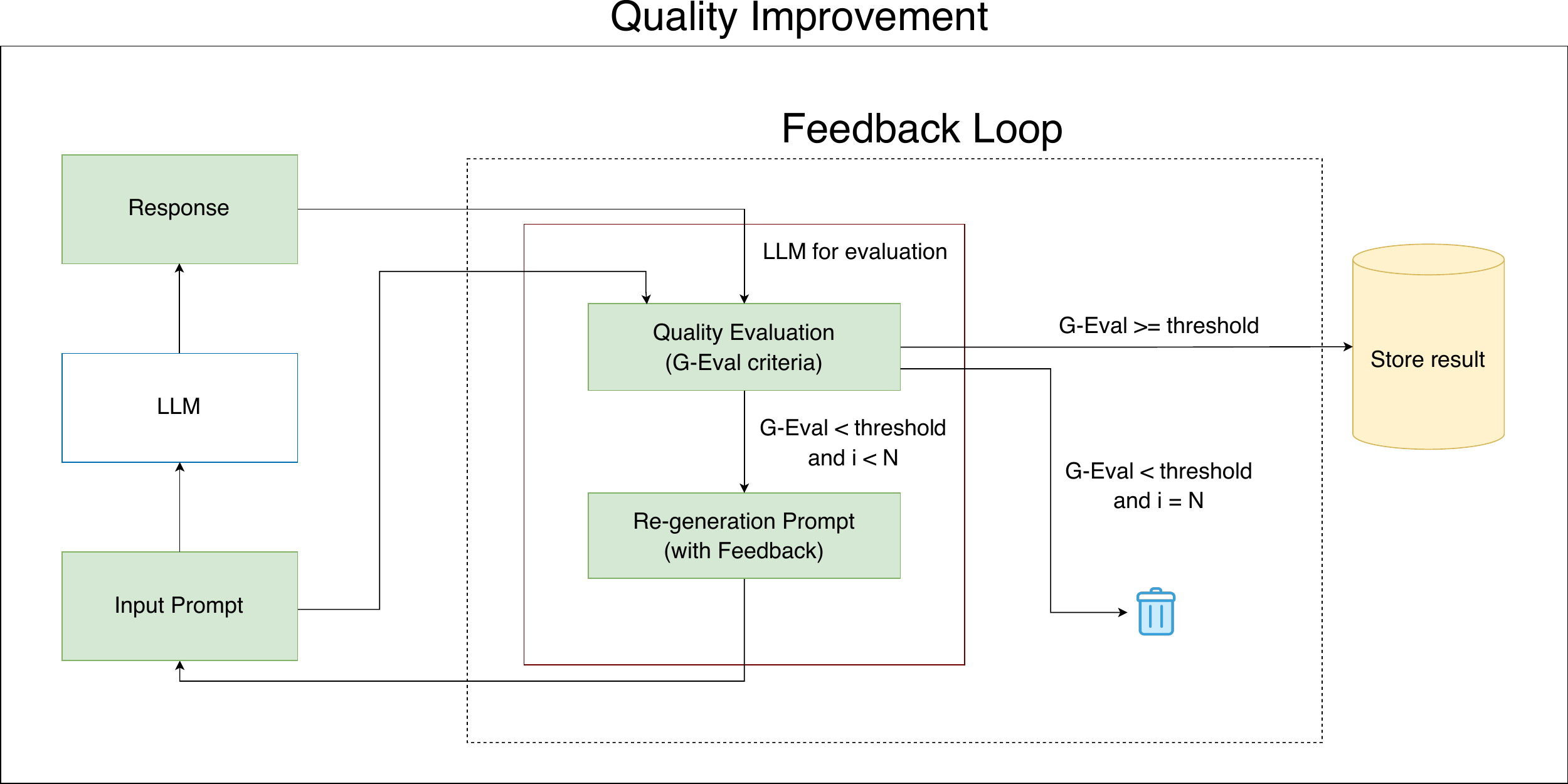}
    \end{center}
    \vspace{-0.1in}
    \caption{Quality Improvement}
    \label{fig:feedback_loop}
\end{figure}

Following the topic generation and filtering phase, the subsequent stage involves instruction generation and quality enhancement.
This is achieved through iterative prompting of an LLM in a process we refer to as the Feedback Loop (Figure~\ref{fig:feedback_loop}).
In each cycle, a random subset of the filtered topics is selected, and the LLM generates a corresponding batch of instructions.
Feeding a random subset of topics aims at increasing the diversity of generated instructions for each iteration.
The feedback loop enables robust filtering and refinement of instruction quality.
This iterative process is essential for mitigating biases, particularly when an auxiliary LLM is used for evaluation, as is the case in our approach~\cite{wang2023selfinstructaligninglanguagemodels}.

Handling the generation of instructions and answers in two separate steps - also defined as "sample-wise decomposition" \cite{long2024llmsdrivensyntheticdatageneration} and supported by \cite{Nad__2025,fan2023chainofthoughttuningmaskedlanguage} - offers flexibility for additional quality control and improvement steps, which reduce potential biases or hallucinations, increasing the overall generation faithfulness; the "sample-wise decomposition" principle is also applied when generating the set of topics before the I\&As themselves.

Once the instruction generation is complete, a dedicated quality improvement phase is applied to ensure only high-quality instructions advance through the pipeline.
As already pointed out, most of the existing approaches do not take any action on low-quality instances, resulting in a potentially drastic reduction of the generated I\&As number.
The exception of Omnieval, which improves medium-quality instances, may not be enough precise compared to a feedback loop approach \cite{long2024llmsdrivensyntheticdatageneration,huang2025datagenunifiedsyntheticdataset}, which is also implemented in our pipeline.
In this approach, the instances deemed low-quality based on evaluation metrics that express important criteria for the evaluation are given to an LLM that provides a feedback to the generation LLM to re-generate the instance.
The process is repeated n times after which, if the instance is not yet improved, it is discarded.
The feedback LLM can be the same or different from the generation LLM, but a different LLM or, more importantly, an LLM from a different family could be more unbiased \cite{panickssery2024llmevaluatorsrecognizefavor,Dai_2024}.

In this work the feedback loop is based on custom G-Eval.
Each instruction is assessed along four key parameters\ -\ \textit{Diversity, Relevance, Conciseness, Correctness} -\ and retained only if it achieves a G-Eval score of at least $\tau$.
The corresponding criterion for each parameter is elaborated in Appendix~\ref{geval_params_criteria}.

The selection of key parameters for instruction quality is grounded in established research across Natural Language Processing (NLP) and Machine Learning.
\textit{Diversity} is essential for mitigating dataset bias and enhancing generalization, as explicit diversity in instruction and data generation has been shown to enhance model robustness and performance \cite{yadav2024explicitdiversityconditionseffective}.
\textit{Relevance} ensures that instructions are contextually appropriate and effective in boosting downstream task performance \cite{kim2024reragimprovingopendomainqa}.
\textit{Conciseness} facilitates comprehension and efficiency.
Wei et al. \cite{wei2022emergentabilitieslargelanguage} demonstrate that clear and concise instructions in prompt design substantially improve the response accuracy and reliability of LLMs.
\textit{Correctness} is also crucial for maintaining the factual and logical integrity of instructions, since inaccuracies can lead to erroneous outputs.
Together, these parameters provide a comprehensive framework for high-quality instruction generation.

\subsubsection{\textbf{Difficulty Enhancement (DFE)}}
\label{dfe}

The Difficulty Enhancement (DFE) module is a critical component within the instruction generation pipeline, tackling the problem of LLM-generated datasets that are too trivial to challenge an evaluated LLM~\cite{patel2025llmgeneratechallengingproblems}.
It consists in paraphrasing instructions and was inspired by the concepts from~\cite{huang2025datagenunifiedsyntheticdataset}. More challenging prompts are necessary to reveal the emergent abilities of LLMs \cite{wei2022emergentabilitieslargelanguage}.

We use a prompt (Appendix~\ref{prompt_for_instructions_difficulty_enhancement}) to direct an LLM to paraphrase instructions\ -\ difficult enough to cause LLMs to answer incorrectly or fail.
The enhanced instructions are then automatically filtered with custom G-Eval, which retains only those meeting the predefined difficulty threshold $\tau$.
This generation procedure of adversarial instructions and filtering enables scalable and consistent dataset curation.
Answers are not considered for DFE as increasing instruction difficulty is sufficient to enhance the overall challenge of the I\&A pair.

\subsubsection{\textbf{Answer Generation and Quality Improvement}}
\label{answer_generation_and_quality_improvement}

The Answer Generation module constitutes a pivotal stage within the data generation pipeline. This process is designed to be modular, thereby facilitating future enhancements such as answer difficulty modulation and the extension to multi-turn dialogue scenarios.

The modularity is achieved by decoupling answer generation from instruction creation. It is crucial for several reasons. First, this sequential approach helps ensure that answers are contextually aligned with high-quality, domain-relevant instructions, thereby reducing the risk of hallucinations and spurious outputs \cite{wang2023selfinstructaligninglanguagemodels}. Moreover, the modular design aligns with the principle of composable pipelines in NLP, which have been shown to support more systematic experimentation and easier integration of novel techniques \cite{liu2021pretrainpromptpredictsystematic}.

For instruction-answer (I\&A) pairs the selected criteria to measure the quality of answers are guided by quality dimensions outlined in~\cite{Fichman2011}. Specifically, we evaluate \textit{Correctness, Relevance, Conciseness, Completeness}, and \textit{Safety}. In comparison to instruction evaluation, we introduce \textit{Safety} as an additional criterion to ensure that answers are free from harmful or biased content, which is essential for upholding ethical standards in AI-generated responses. We also include \textit{Completeness} to determine whether answers thoroughly address both explicit and implicit requirements of the query.

The feedback loop in answer improvement is similar to the process in instruction improvement (Figure~\ref{fig:feedback_loop}), but with different evaluation criteria (Appendix \ref{geval_params_criteria_answer}). The improved instruction-answer pairs are subsequently forwarded to next processing phase for further diversity enhancement.

\subsubsection{\textbf{Diversity Enhancement (DVE)}}
\label{dve}

The diversity enhancement (DVE) plays an important role in enhancing the diversity and quality of the generated I\&A pairs within the synthetic data generation pipeline.
In this stage, embeddings for each I\&A pair are computed using a multilingual embedding model, which captures both semantic and linguistic features across languages.
Subsequently, the DVE algorithm is applied to filter the pairs based on their embedding-space distances.
Specifically, the algorithm iterates through the set of I\&A pairs, retaining only those that are sufficiently dissimilar, similar to what was done by~\cite{huang2025datagenunifiedsyntheticdataset}.
This ensures the final dataset contains maximally diverse examples, both in phrasing and underlying meaning.
The steps are as follows:
\begin{itemize}
    \item \textbf{Embedding Calculation:} Use a multi-lingual embedding model to generate embeddings for each I\&A item.
    \item \textbf{Pairwise Distance Computation:} Compute pairwise distances between I\&A embeddings within each group.
    \item \textbf{Threshold Filtering:} For each I\&A, compare its distance to other I\&As in the group. If the minimum pairwise distance between an I\&A and any other I\&A in the group is less than a specified threshold, discard that I\&A to remove redundancy.
    \item \textbf{DVE Output:} Collect all remaining I\&As that pass the distance threshold filtering into a DVE pool.
\end{itemize}

Embedding-based deduplication and diversity filtering have become standard practice to enhance dataset variety while mitigating redundancy. Wang et al. \cite{wang2023selfinstructaligninglanguagemodels} demonstrated that leveraging embedding distances for deduplication maximizes both linguistic and semantic diversity in synthetic corpora.
Redundancy in synthetic datasets not only inflates data size but also introduces biases and reduces the generalization capacity of models trained on such data \cite{Honovich2023}, as evidenced in popular datasets such as Stanford Alpaca \cite{alpaca}.
By incorporating embedding-based filtering, the pipeline addresses these challenges directly, ensuring that only distinct I\&A pairs contribute to downstream learning.

\subsubsection{Generation Models Overview}

In our pipeline, we employ a suite of LLMs and embedding model(s), as detailed in Table~\ref{tab:pipeline-stages-models}.
The Gemini-1.5-pro-002 model is primarily used for generation tasks due to its strong generative capabilities and nuanced understanding of instructional content, while Gemini-2.0-flash-001 was leveraged for filtering and evaluation phases, owing to its efficiency and robustness in discerning quality and relevance.
For DVE, we integrate the bge-m3 embedding model, which excels in producing high-quality semantic representations, thereby enabling effective grouping and duplicate detection within the synthetic dataset.

\begin{table}[!h]
    \centering
    \small
    \setlength{\tabcolsep}{6pt}%
    \renewcommand{\arraystretch}{0.95}%
    \caption{Pipeline Stages and Models}
    \label{tab:pipeline-stages-models}
    \begin{tabular}{lll}
        \toprule
        \bfseries Phase & \bfseries Stage & \bfseries Model \\
        \midrule
        \multirow{2}{*}{Topic}
        & generation & gemini-1.5-pro-002 \\
        & filtering  & gemini-2.0-flash-001 \\
        \addlinespace
        \multirow{3}{*}{Instruction}
        & generation & gemini-1.5-pro-002 \\
        & filtering  & gemini-2.0-flash-001 \\
        & evaluation & gemini-2.0-flash-001 \\
        \addlinespace
        \multirow{2}{*}{DFE}
        & generation & gemini-1.5-pro-002 \\
        & filtering  & gemini-2.0-flash-001 \\
        \addlinespace
        \multirow{3}{*}{Answer}
        & generation & gemini-1.5-pro-002 \\
        & filtering  & gemini-2.0-flash-001 \\
        & evaluation & gemini-2.0-flash-001 \\
        \addlinespace
        DVE &            & bge-m3 \\
        \bottomrule
    \end{tabular}
\end{table}

\subsection{Evaluation Pipeline}
Our system includes an evaluation pipeline designed to quantify model performance.
This component is what makes the system truly end-to-end, enabling full benchmarking cycles from generation to evaluation.

In this study its primary role is to validate the data generation method: by evaluating on a human-curated dataset and its synthetic equivalent, the pipeline provides the performance scores needed to compute the gap between them.
The validation principle is to determine how closely the synthetic dataset replicates the original benchmark.
A smaller distance indicates that the synthetic dataset is a high-fidelity substitute for the original, thus confirming the success of the generation pipeline.

\section{Experiment Setup}
\label{experimental_setup_and_model_selection}

Experiments were designed to assess the fidelity of the synthetic data generation pipeline.
This was done by comparing model performance on generated datasets against performance on a human-curated benchmark.

\subsection{Datasets and Scope}

\begin{figure}[htbp]
    \begin{center}
        \includegraphics[width=0.5\linewidth]{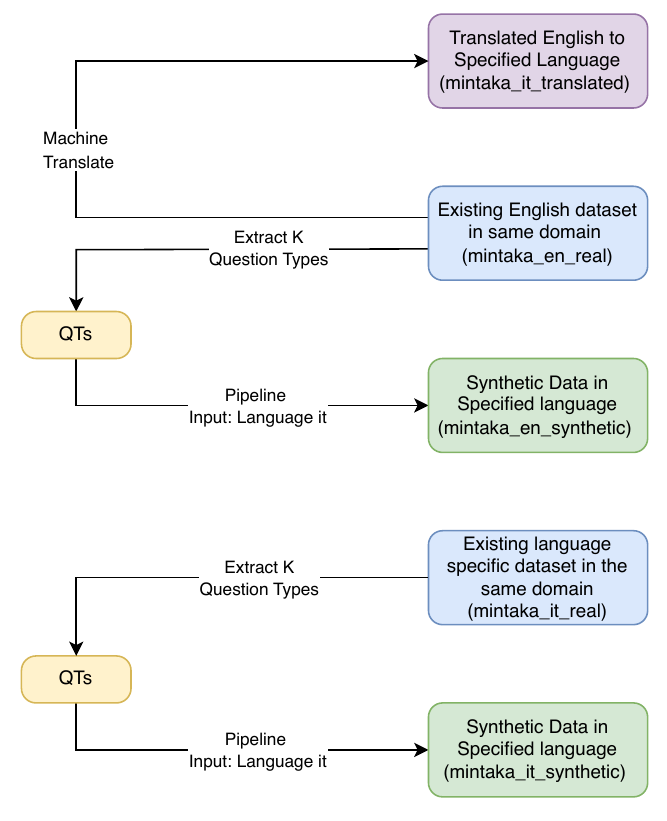}
    \end{center}
    \vspace{-0.1in}
    \caption{Language Datasets Diagram}
    \label{fig:language}
\end{figure}

To evaluate the system's ability to generate language-specific datasets, we used the professionally translated Mintaka dataset as our ground-truth benchmark~\cite{sen2022mintakacomplexnaturalmultilingual}.
Our experiment involved datasets in both English and Italian, structured as follows (see Figure~\ref{fig:language}):
\textbf{Original Datasets:} The original Mintaka question-answer pairs in English and Italian.
\textbf{Machine-Translated Dataset:} A control set created by translating the Original Mintaka English set to Italian using an automated API, which helps benchmarking the quality of standard translation methods.
\textbf{Synthetic Datasets:} Datasets generated by our pipeline in English and Italian, using 8 Question Templates (QTs) designed to match the domains in Mintaka.

For the final evaluation, all datasets were reduced by randomly sampling 1,500 pairs from each.
This ensures a fair comparison across datasets of equal size while maintaining manageable computation times.

\subsection{Data Generation Parameters}

The synthetic datasets were created using the parameters detailed in Table~\ref{tab:generation-parameters}.
These values were chosen to balance instruction diversity with topic coherence, avoiding the repetition often observed with larger generation batches.
They allowed to generate, for each run of the generation pipeline, around 2,500 I\&A pairs.
The DVE similarity threshold is always 0.3, as higher thresholds were not providing significant improvements in diversity, while reducing the dataset size too much.

\begin{table}[!h]
    \centering
    \small
    \setlength{\tabcolsep}{6pt}%
    \renewcommand{\arraystretch}{0.95}%
    \caption{Generation Pipeline Parameters}
    \label{tab:generation-parameters}
    \begin{tabular}{lr}
        \toprule
        Parameter & Value \\
        \midrule
        G-Eval feedback threshold \(\tau\)      & 8   \\
        Number of QTs                           & 8   \\
        Number of topics                        & 20  \\
        Number of sampled topics                & 5   \\
        Number of iterations                    & 50  \\
        Number of instructions per iteration    & 50  \\
        DVE similarity threshold                & 0.3 \\
        \bottomrule
    \end{tabular}
\end{table}

\subsection{Evaluated Models}

We evaluated two distinct models to ensure the robustness of our findings across different levels of capability:
\textbf{A strong model:} Gemini 2.5 Flash (2025), a high-performing LLM with an 83.6\% score on MMLU-pro.
\textbf{A weak model:} Llama 2 Chat 13B (2023), a smaller LLM with a 25.3\% score on MMLU-pro.
This choice of diverse models helps mitigate the risk of self-enhancement bias, where an LLM may unfairly favor outputs generated by itself or a closely related model~\cite{chen2025llmevaluatorspreferreason}.

\subsection{Evaluation Metrics}

We employed a suite of metrics to provide a comprehensive assessment of performance:
\textbf{ROUGE-L:} To measure lexical overlap, effective for short answers.
\textbf{BertScore:} To measure semantic similarity, crucial for multilingual contexts.
\textbf{Answer Relevance:} To measure how well answers address the given questions, penalizing incompleteness or redundancy \cite{es-etal-2024-ragas}.
\textbf{Custom G-Eval:} Our LLM-as-a-judge metric for assessing overall quality, as described in Section~\ref{custom_geval}.
The criteria chosen for this case were \textit{Accuracy, Relevance and Completeness}: the last two are detailed in Sections~\ref{instruction_generation_and_quality_improvement} and~\ref{answer_generation_and_quality_improvement}, while \textit{Accuracy} has been introduced to explicitly compare the obtained answer with the golden one.
The model chosen as judge was Gemini 2.5 Pro.

Together, these metrics provide a nuanced view of the quality and fidelity of the generated datasets.

\section{Results}
\label{result}

\begin{table*}[t]
    \caption{Experiments Results on Strong Model}
    \label{tab:strong-model-evaluation-1}
    \vskip 0.1in
    \centering
    \begin{small}
        \begin{sc}
            \resizebox{\textwidth}{!}{%
                \begin{tabular}{lll|cccc|ccc}
                    \toprule
                    \bf Dataset & \bf DVE & \bf DFE & \bf G-Eval \textbar  $\Delta$ & \bf Acc. \textbar $\Delta$ & \bf Rel. \textbar $\Delta$ & \bf Compl. \textbar $\Delta$ & \bf Rouge-L \textbar $\Delta$ & \bf BScore F1 \textbar $\Delta$ & \bf AR \textbar $\Delta$  \\
                    \midrule
                    Real en & - & - & 8.17 & 8.665 & 8.476 & 8.533 & 0.397 & 0.547 & 0.523  \\
                    Synthetic en & No & No & 9.43 \textbar +12.6\% & 9.53 \textbar +8.7\% & 9.32 \textbar +8.4\% & 9.42 \textbar +8.9\% & 0.21 \textbar -18.7\% & 0.555 \textbar +0.8\% & 0.793 \textbar +13.5\%  \\
                    Synthetic en & Yes & No & 9.14 \textbar +9.7\% & 9.327 \textbar +6.6\% & 9.101 \textbar +6.3\% & 9.005 \textbar +4.7\% & 0.316 \textbar -8.1\% & 0.604 \textbar +5.7\% & 0.781 \textbar +12.9\% \\
                    Synthetic en & Yes & Yes & 8.74 \textbar +5.7\% & 9.157 \textbar +4.9\% & 8.331 \textbar -1.4\% & 8.737 \textbar +2.0\% & 0.086 \textbar -31.1\% & 0.474 \textbar -7.3\% & 0.782 \textbar +13.0\% \\ \midrule
                    Real it & - & - & 8.00 & 8.141 & 7.846 & 8.007 & 0.253 & 0.447 & 0.405 \\
                    Translated it & - & - & 8.23 \textbar +2.3\% & 8.338 \textbar +2.0\% & 8.176 \textbar +3.3\% & 8.189 \textbar +1.8\% & 0.273 \textbar +2.0\% & 0.478 \textbar +3.1\% & 0.38 \textbar -1.2\% \\
                    Synthetic it & No & No & 8.91 \textbar +9.1\% & 9.091 \textbar +9.5\% & 8.741 \textbar +9.0\% & 8.913 \textbar +9.1\% & 0.198 \textbar -5.5\% & 0.606 \textbar +15.9\% & 0.724 \textbar +16.0\% \\
                    Synthetic it & Yes & No & 8.86 \textbar +8.6\% & 9.057 \textbar +9.2\% & 8.647 \textbar +8.0\% & 8.888 \textbar +8.8\% & 0.166 \textbar -8.7\% & 0.558 \textbar +11.1\% & 0.67 \textbar +13.2\% \\
                    Synthetic it & Yes & Yes & 8.58 \textbar +5.8\% & 8.681 \textbar +5.4\% & 8.469 \textbar +6.2\% & 8.593 \textbar +5.9\% & 0.114 \textbar -13.9\% & 0.586 \textbar +13.9\% & 0.731 \textbar +16.3\%  \\
                    \bottomrule
                \end{tabular}
            }
        \end{sc}
    \end{small}
    \vskip 0.1in
\end{table*}

Starting from the strong model evaluations (Table~\ref{tab:strong-model-evaluation-1}), some trends emerge.
The base synthetic English dataset inflates the average G-Eval score by +12.6\% compared to the real English Mintaka.
The DVE reduces this inflation, but DVE \& DFE manage to lower the gap to just +5.7\%.
When breaking down G-Eval into components all synthetic datasets present higher Accuracy and Completeness than the real English Mintaka, suggesting that the pipeline, even with enhancements, still generates instructions that may be too simple for evaluated LLMs to answer correctly and comprehensively.
However, the average Relevance has a significant drop on the DVE \& DFE dataset (-7.7\% compared to DVE only), bringing it closer to the real one and proving that these enhancements are effective to generate data that tests the model's ability to stay on topic.
As for the other metrics, on BertScore F1 the DVE \& DFE dataset is the only one that drops below the real dataset's score, which indicates that the model may need to use the language in a way that is semantically different from the golden answer; the Answer Relevance metric stays high on all datasets, with the real-world data outperforming all synthetic counterparts; finally, the Rouge-L score plummets on the DVE \& DFE dataset, which suggests that on this dataset the model paraphrases heavily and generates answers that are syntactically very different from the golden answer.

On the Italian experiments the patterns are consistent, showing that the DVE \& DFE dataset brings the average G-Eval score closest to the real one (+5.8\% difference).
As for the G-Eval components breakdown the DVE \& DFE dataset drops evenly on all of them compared to the other synthetic versions.
Rouge-L again drops to the lowest value among all the Italian datasets.
The translated Mintaka dataset is slightly easier than the real one on average G-Eval (+2.3\%), suggesting that the translation may have removed linguistic ambiguity or complex expressions present in the real counterpart, making the tasks less challenging.

\begin{table*}[t]
    \caption{Experiments Results on Weak Model}
    \label{tab:weak-model-evaluation-1}
    \vskip 0.1in
    \centering
    \begin{small}
        \begin{sc}
            \resizebox{\textwidth}{!}{%
                \begin{tabular}{lll|cccc|ccc}
                    \toprule
                    \bf Dataset & \bf DVE & \bf DFE & \bf G-Eval \textbar $\Delta$ & \bf Acc. \textbar $\Delta$ & \bf Rel. \textbar $\Delta$ & \bf Compl. \textbar $\Delta$ & \bf Rouge-L \textbar $\Delta$ & \bf BScore F1 \textbar $\Delta$ & \bf ARel. \textbar $\Delta$ \\
                    \midrule
                    Real en & - & - & 5.69 & 5.683 & 5.690 & 5.700 & 0.169 & 0.389 & 0.233 \\
                    Synthetic en & Yes & No & 7.33 \textbar +16.4\% & 7.272 \textbar +15.9\% & 7.363 \textbar +16.7\% & 7.359 \textbar +16.6\% & 0.203 \textbar +3.4\% & 0.569 \textbar +18.0\% & 0.690 \textbar +22.9\% \\
                    Synthetic en & Yes & Yes & 6.78 \textbar +10.9\% & 6.562 \textbar +8.8\% & 6.942 \textbar +12.5\% & 6.825 \textbar +11.3\% & 0.134 \textbar -3.5\% & 0.543 \textbar +15.4\% & 0.707 \textbar +23.7\% \\ \midrule
                    Real it & - & - & 4.28 & 4.464 & 4.031 & 4.336 & 0.160 & 0.401 & 0.242 \\
                    Translated it & - & - & 4.20 \textbar -0.8\% & 4.206 \textbar -2.6\% & 4.237 \textbar +2.1\% & 4.169 \textbar -1.7\% & 0.148 \textbar -1.2\% & 0.411 \textbar +1.0\% & 0.310 \textbar +3.4\% \\
                    Synthetic it & Yes & No & 4.51 \textbar +2.3\% & 4.582 \textbar +1.2\% & 4.386 \textbar +3.6\% & 4.571 \textbar +2.4\% & 0.163 \textbar +0.3\% & 0.577 \textbar +17.6\% & 0.511 \textbar +13.5\% \\
                    Synthetic it & Yes & Yes & 4.35 \textbar +0.7\% & 4.113 \textbar -3.5\% & 4.396 \textbar +3.7\% & 4.546 \textbar +2.1\% & 0.139 \textbar -2.1\% & 0.593 \textbar +19.2\% & 0.566 \textbar +16.2\% \\
                    \bottomrule
                \end{tabular}
            }
        \end{sc}
    \end{small}
    \vskip 0.1in
\end{table*}

Moving onto the weak model experiments (Table~\ref{tab:weak-model-evaluation-1}), the trends are overall confirmed, but the synthetic datasets have a bigger G-Eval inflation from the real ones compared to the strong model experiments.
In fact English DVE \& DFE is the synthetic dataset with average G-Eval closest to the real one (+10.9\%), but this distance is double compared to the strong model evaluation results, signaling that the weak model is disproportionately helped by this synthetic dataset, while being "punished" by the real one.

BertScore F1 and Answer Relevance on English synthetic datasets are much higher than the real dataset ones, suggesting that the real dataset is more complex and has a more difficult I\&A structure for weak models, while the synthetic datasets may contain structural cues or keywords that the LLM can use to generate a semantically similar answer.
Rouge-L is again the lowest on English DVE \& DFE dataset against all the English ones.

In Italian, the DVE \& DFE synthetic dataset is almost a perfect overlap with the real one in terms of average G-Eval (+0.7\%).
It is interesting to notice that, while the overall G-Eval is almost the same, the synthetic dataset obtains the lowest results on Accuracy, whereas the real one on Relevance, suggesting that even the best synthetic result may not perfectly replicate the reason for models' low benchmark.
The conclusions on all other metrics are the same as on the English weak model experiments.
The translated Mintaka dataset is slightly harder for the weak model compared to the real one, as this LLM is tied to the specific statistical patterns that it was trained on and the unnatural artifacts introduced by the translation may represent an out-of-distribution input \cite{yang2024adversarialrobustnessoutofdistributionrobustness,yuan2023revisitingoutofdistributionrobustnessnlp}.

\section{Discussion and limitations}
\label{discussion}

The implemented pipeline is simple to configure, offers the expected customizability and the DVE/DFE mechanisms prove effective according to the experiments.
Overall, by averaging the G-Eval scores of strong and weak LLMs evaluations on both English and Italian datasets, the distance between the DVE \& DFE datasets and the original Mintaka is +5.7\%.
This result demonstrates that the system is capable of benchmarking LLM-based applications by generating high-quality medium-complexity evaluation datasets with custom language and content, possibly introducing a new flexible monitoring paradigm for LLMOps.
The biggest delta for the DVE \& DFE datasets is measured when evaluating the weak model in English (+10.9\%).
This suggests that while the synthetic data is a good substitute for human-curated data in many cases, it may contain artifacts or patterns that smaller models exploit, or its complexity may not fully challenge them in the same way as human-curated data.

The scope of this study presents clear directions for future improvements.
First, the meta-evaluation should be expanded to a larger cohort of models for a more comprehensive comparison, including: (a) using models from different families to better mitigate self-enhancement bias, and (b) exploring the use of generator/judge ensembles to potentially enhance overall quality.

Additionally, the current meta-evaluation is based on a single benchmark dataset (Mintaka).
A necessary next step is to validate the pipeline against a wider variety of datasets, which could include a domain-specific benchmark for a fine-tuned LLM, fully assessing the system's capabilities in specialized contexts.

A key finding was that the machine-translated Italian dataset performed better than the synthetic Italian DVE \& DFE dataset.
Future work should include a human evaluation with native speakers on a small, randomly sampled sets of instruction-answer pairs, to analyze the cultural specificity of all three Italian datasets and determine the overall value of the language-specific feature.

Finally, the human evaluation proposed for the Italian datasets could be extended more broadly.
Since our system uses LLMs in every phase, it risks introducing subtle, systemic biases that automated metrics might miss.
Applying targeted human evaluation is therefore a crucial step in highlighting these potential pipeline-induced biases.

\section{Conclusion}
\label{conclusion}

This research aimed to show if it is possible to automate the evaluation of specialized LLM-based applications via synthetic textual benchmarks.
The goal was tackled by leveraging the generation capabilities of modern LLMs and applying rigorous control mechanisms, resulting in a system that not only generates datasets with tens of thousands of I\&A pairs, but also provides a way to apply them to benchmark LLM-based applications.
The benchmarking data generated by the pipeline proved to be as effective as medium-difficulty benchmarking datasets, both in English and Italian.

On top of this research a further assessment of its capabilities regarding domain specificity, cultural specificity and more generation, filtering or evaluated LLMs, may open to new uses and improvements.
Given the relevance of RAG applications, the pipeline can be extended for their evaluation by generating synthetic source documents and I\&As grounded on them.
Moreover, the components of the whole system are independent, meaning that they can be re-used as needed. For instance, the generation pipeline has the potential to generate datasets for training or fine-tuning LLMs via reinforcement learning, but only by considering the risk of dataset contamination.

\bibliographystyle{plainnat}

\clearpage
\appendix

\section{G-Eval Prompts}

\subsection{Prompt for Topics Generation}
\label{prompt_for_topics_generation}

\begin{lstlisting}[breaklines=true, basicstyle=\scriptsize\ttfamily, frame=single]
p_template = """<|start_header_id|>system<|end_header_id|>Please follow my instruction very carefully to generate 20 diverse topics for a specific question type.
Here are the requirements:
1. Try not to repeat the words for each topic to maximize diversity.
2. Each topic must contain three words maximum.
3. Topics are not questions, just general topics.
4. Topics must be 20 in total for each question type, always.
5. Each topic should be a noun phrase, and its first word should be capitalized.
6. The topics should be closely related to the given question type.
7. Output your answer in json format like this:
    {{
    "topics": A JSON list of 20 topics related to the given question type, following the requirements
}}<|eot_id|>
8. The list of topics must be a JSON list, not surrounded by quotes, just by square brackets.
<|start_header_id|>user<|end_header_id|>The question type is: {question_type}<|eot_id|>
<|start_header_id|>assistant<|end_header_id|>{format_instructions}. Json Output:"""
\end{lstlisting}

\subsection{Auxiliary Prompt for Topic Evaluation When Filtering Topics}
\label{auxiliary_prompt_for_topic_evaluation_when_filtering_topics}

\begin{lstlisting}[breaklines=true, basicstyle=\scriptsize\ttfamily, frame=single]
p_topic_template_no_formatting = """<|start_header_id|>system<|end_header_id|>Please follow my instruction very carefully to generate 1 topic for a specific question type.
Here are the requirements:
1. The topic must contain three words maximum.
2. Topics are not questions, just general topics.
3. A topic should be a noun phrase, and its first word should be capitalized.
4. The topic should be closely related to the given question type.
5. Output your answer in json format like this:
    {{
    "topic": One topic related to the given question type, following the requirements
}}<|eot_id|>
<|start_header_id|>user<|end_header_id|>The question type is: {question_type}<|eot_id|>
<|start_header_id|>assistant<|end_header_id|>. Json Output:
"""
\end{lstlisting}

\subsection{Prompt for All Stages of Evaluation}
\label{prompt_for_all_stages_of_evaluation}

\begin{lstlisting}[breaklines=true, basicstyle=\scriptsize\ttfamily, frame=single]
p_template_for_evaluation = """<|start_header_id|>system<|end_header_id|>
Given the evaluation criteria below which outlines how you should judge a conversation between a user and an LLM chatbot using the {parameters} fields in each turn, return only a JSON object with two keys:
- "score": an integer from 0 to 10
- "reason": a concise justification (do not quote the score in the reason)

Please mention specific information from {parameters} in the conversation in your reason.
Do not include any explanation, schema, or extra text. Output only the JSON object.

Evaluation Criteria:
    {criteria}

Example:
    {{
    "score": 0,
    "reason": "The text does not follow the criteria provided."
}}

**
IMPORTANT: When scoring, penalize heavily those results that are not accurate enough based on the provided criteria, giving very low scores in that case.
We want to achieve the best quality results. The score you output will be averaged, so in case it is low, it must be very low to move the average.
IMPORTANT: Please make sure to only return in JSON format, with the "score" and "reason" key. No words or explanation is needed.
IMPORTANT: If the user requested that the output must not be longer than a maximum number of words, it means that responses with more than this number of words don't follow the criteria
(e.g. for "Output a maximum of 2 words", responses with three and more words don't follow the criteria).

<|eot_id|>
<|start_header_id|>conversation_to_evaluate<|end_header_id|>
    {user_prompt}
<|eot_id|>
<|start_header_id|>assistant<|end_header_id|>
    {assistant_response}
<|eot_id|>
<|eot_id|>
JSON:"""
\end{lstlisting}

\subsection{Prompt for Initial Batch Instructions Generation}
\label{prompt_for_initial_batch_instructions_generation}

\begin{lstlisting}[breaklines=true, basicstyle=\scriptsize\ttfamily, frame=single]
p_template_for_instruction_generation = """<|start_header_id|>system<|end_header_id|>
You are asked to come up with a set of {number_of_instructions} diverse instructions that a machine learning model can't answer, or will answer with the wrong facts, related to one or more topics from the given list.
The topics have been derived from a specific instruction domain, which is also given.

Here are the requirements:
1. Try not to repeat the words for each instruction to maximize diversity.
2. The language used for the instruction also should be diverse. For example, you should combine questions with imperative instructions.
3. The type of instructions should be diverse. The set should include diverse types of instructions, such as:
    {instruction_types}
4. The instructions should be in {instruction_language}.
5. Each instruction should be short and concise, as a single sentence. Either an imperative sentence or a question is permitted.
6. Every quote inside each instruction should be single-quoted, not double-quoted. Do not escape single quotes inside the instruction.
7. I will give you instruction domain and topics to help you brainstorm the instructions.
8. Output your answer in JSON format like this:
    {{
    "instructions": A JSON list of {number_of_instructions} instructions related to the given topics, following the requirements
}}<|eot_id|>
<|start_header_id|>user<|end_header_id|>
The topics are: {topics}
The instruction domain is: {instruction_domain}
<|eot_id|>
<|start_header_id|>assistant<|end_header_id|>{format_instructions}. Json Output:
"""
\end{lstlisting}

\subsection{Auxiliary Prompt for Instruction Evaluation When Filtering Instructions}
\label{auxiliary_prompt_for_instruction_evaluation_when_filtering_instructions}

\begin{lstlisting}[breaklines=true, basicstyle=\scriptsize\ttfamily, frame=single]
p_template_for_single_instruction_generation = """<|start_header_id|>system<|end_header_id|>
You are asked to come up with a set of 1 instruction that a machine learning model can't answer, or will answer with the wrong facts, related to one or more topics from the given list.
The topics have been derived from a specific instruction domain, which is also given.

Here are the requirements:
1. Syntactically speaking, the instruction can either be a question or imperative instructions.
2. The instruction can fall in one of these types:
    {instruction_types}
3. The instruction should be in {instruction_language}.
4. The instruction should be short and concise, as a single sentence. Either an imperative sentence or a question is permitted.
5. I will give you instruction domain and topics to help you brainstorm the instructions.
6. Every quote inside each instruction should be single-quoted, not double-quoted. Do not escape single quotes inside the instruction.
7. Output your answer in JSON format like this:
    {{
    "instruction": One instruction related to the given topics, following the requirements
}}<|eot_id|>
<|start_header_id|>user<|end_header_id|>
The topics are: {topics}
The instruction domain is: {instruction_domain}
<|eot_id|>
<|start_header_id|>assistant<|end_header_id|> Json Output:
"""
\end{lstlisting}

\subsection{Prompt for Batch Instructions Improvement}
\label{prompt_for_batch_instructions_improvement}

\begin{lstlisting}[breaklines=true, basicstyle=\scriptsize\ttfamily, frame=single]
p_template_for_instructions_improvement = """<|start_header_id|>system<|end_header_id|>
You are asked to improve a set of instructions that a machine learning model can't answer, or will answer with the wrong facts, related to one or more topics from the given list.
The topics have been derived from a specific instruction domain, which is also given.

Here are the requirements:
1. Syntactically speaking, the instructions can either be a question or imperative instructions.
2. The instruction can fall in one of these types:
    {instruction_types}
3. The instructions should be in {instruction_language}.
4. The instructions should be short and concise, as a single sentence. Either an imperative sentence or a question is permitted.
5. I will give you instructions domain and topics to help you improve the instructions.
6. Do not output more instructions than the provided ones. Just improve the provided ones.
7. Output your answer in JSON format like this:
    {{
    "instructions": A JSON list of improved instructions related to the given topics, following the requirements
}}<|eot_id|>
The topics are: {topics}
The instruction domain is: {instruction_domain}
The instructions to improve are: {instructions}
<|eot_id|>
Json Output:
"""
\end{lstlisting}

\subsection{Prompt for Instructions Difficulty Enhancement}
\label{prompt_for_instructions_difficulty_enhancement}

\begin{lstlisting}[breaklines=true, basicstyle=\scriptsize\ttfamily, frame=single]
p_template_for_instructions_difficulty_enhancement = """<|start_header_id|>system<|end_header_id|>
You are asked to improve the difficulty of a set of existing instructions.
A machine learning model or LLM should not be able to answer these instructions, or will answer with the wrong facts.
We say that these instructions should be adversarial to the model.

Here are the requirements:
1. The difficulty of the instructions should be improved in one or more of the following ways:
- Paraphrase the instructions to make them more complex or challenging.
- Introduce ambiguity or multiple interpretations to the instructions to make them more difficult.
2. When the instructions have multiple choices, you must also improve the difficulty of the choices in one or more of the following ways:
- Paraphrase the choices to make them more complex or challenging.
- Add a new plausible choice to the existing ones, which is not the correct answer.
3. You don't change the content or language of the instructions, just improve their difficulty.
4. The instructions should be short and concise, as a single sentence.
5. Syntactically speaking, the instructions can either be questions or imperative instructions.
6. Do not output more instructions than the provided ones. Just improve the provided ones.
7. The instructions should be in {instruction_language}.
8. Output your answer in JSON format like this:
    {{
    "instructions": A JSON list of difficulty improved instructions, following the requirements
}}<|eot_id|>
The instructions to improve are: {instructions}
<|eot_id|>
Json Output:
"""
\end{lstlisting}

\subsection{Prompt for Single Answer Generation}
\label{prompt_for_single_answer_generation}

\begin{lstlisting}[breaklines=true, basicstyle=\scriptsize\ttfamily, frame=single]
p_template_for_answers_generation = """<|start_header_id|>system<|end_header_id|>
You are asked to output an answer to a given instruction.

Here are the requirements:
1. The answer must be semantically correct for the given instruction.
2. The answer must be syntactically correct for the given instruction.
3. In case the instructions ask about something personal, simply state that you don't know the answer.
4. Provide a concise and accurate answer to the instruction. Avoid verbose details and keep the response limited to one short sentence or a few factual words.
5. The answer should be in {instruction_language}.
6. Every quote inside each answer should be single-quoted, not double-quoted. Do not escape single quotes inside the instruction.
7. Output your answer in JSON format like this:
    {{
    "answer": One answer to the given instruction, following the requirements
}}<|eot_id|>
<|start_header_id|>user<|end_header_id|>
The instruction is: {instruction}
<|eot_id|>
<|start_header_id|>assistant<|end_header_id|>. Json Output:
"""
\end{lstlisting}

\subsection{Prompt for Single Answer Improvement}
\label{prompt_for_single_answer_improvement}

\begin{lstlisting}[breaklines=true, basicstyle=\scriptsize\ttfamily, frame=single]
p_template_for_answers_improvement = """<|start_header_id|>system<|end_header_id|>
You are asked to improve 1 answer which is generated from its corresponding instruction.

Here are the improvement strategies:
1. Improve semantic correctness: Ensure the answer accurately addresses the original instruction
2. Improve syntactic correctness: Fix grammatical errors and structure issues
3. Personal queries: If an instruction asks about personal experiences, output "I don't know"
4. The answer should be in {instruction_language}.
5. Conciseness: Remove unnecessary verbosity while preserving key information. Limit to one very short sentence or few factual words, if possible
6. Bias awareness: Present balanced perspectives on controversial topics
7. Safety compliance: Reject harmful requests with ethical compliance notice
8. Output your answer in JSON format like this:
    {{
    "answer": An answer related to their instruction, following the requirements
}}<|eot_id|>
The instruction is: {instruction}
The answer to improve is: {answer}
<|eot_id|>
Json Output:
"""
\end{lstlisting}

\section{G-Eval Parameters and Criteria}
\label{geval_params_criteria}

\subsection{Topic Evaluation}

\begin{lstlisting}[breaklines=true, basicstyle=\scriptsize\ttfamily, frame=single]
PARAMETERS=[["Relatedness"], ["Correctness"]]
CRITERIA=["Relatedness to the question type", "Semantic correctness"]
\end{lstlisting}

\subsection{Instruction Evaluation}

\begin{lstlisting}[breaklines=true, basicstyle=\scriptsize\ttfamily, frame=single]
PARAMETERS = [["Diversity"], ["Relevance"], ["Conciseness"], ["Correctness"]]
CRITERIA = [
"Instructions use diverse language and types.",
"Instructions are relevant to the given topics and instruction domain.",
f"Each instruction is concise, a single sentence, and in {instruction_language}.",
"Instructions are syntactically and semantically correct."
]
\end{lstlisting}

\subsection{Difficulty Enhancement}

\begin{lstlisting}[breaklines=true, basicstyle=\scriptsize\ttfamily, frame=single]
DIFFICULTY_ENHANCED_PARAMETERS = [["Difficulty"], ["Conciseness"], ["Correctness"]]
DIFFICULTY_ENHANCED_CRITERIA = [
"Each instruction is adversarial for a classical model or LLM.",
f"Each instruction is concise, a single sentence, and in {instruction_language}.",
"Instructions are syntactically and semantically correct."
]
\end{lstlisting}

\subsection{Answer Evaluation}
\label{geval_params_criteria_answer}

\begin{lstlisting}[breaklines=true, basicstyle=\scriptsize\ttfamily, frame=single]
PARAMETERS = [["Correctness"], ["Relevance"], ["Conciseness"], ["Completeness"], ["Safety"]]
CRITERIA = [
"Answer is factually accurate and logically sound. Contains no factual errors or contradictions.",
"Answer directly addresses all instruction aspects, no tangents.",
"Answer avoids repetition and irrelevant details.",
"Answer fully resolves explicit/implicit query requirements",
"Personal queries receive 'I don't know' responses without elaboration. Harmful/illegal requests trigger ethical refusal without providing alternatives."
]
\end{lstlisting}

\subsection{Meta-Evaluation}
\label{geval_params_meta}

\begin{lstlisting}[breaklines=true, basicstyle=\scriptsize\ttfamily, frame=single]
PARAMETERS = [["Accuracy"], ["Relevance"], ["Completeness"]]
CRITERIA = [
"Answer is accurate in relation to the expected answer. Contains no factual errors or contradictions.",
"Answer directly addresses all instruction aspects according to the expected answer, no tangents.",
"Answer fully resolves explicit/implicit query requirements according to the expected answer",
]
\end{lstlisting}

\end{document}